# Multiresolution Tree Networks for 3D Point Cloud Processing

Matheus Gadelha, Rui Wang and Subhransu Maji

College of Information and Computer Sciences,
University of Massachusetts - Amherst
{mgadelha, ruiwang, smaji}@cs.umass.edu

**Abstract.** We present multiresolution tree-structured networks to process point clouds for 3D shape understanding and generation tasks. Our network represents a 3D shape as a set of locality-preserving 1D ordered list of points at multiple resolutions. This allows efficient feed-forward processing through 1D convolutions, coarse-to-fine analysis through a multi-grid architecture, and it leads to faster convergence and small memory footprint during training. The proposed tree-structured encoders can be used to classify shapes and outperform existing point-based architectures on shape classification benchmarks, while tree-structured decoders can be used for generating point clouds directly and they outperform existing approaches for image-to-shape inference tasks learned using the ShapeNet dataset. Our model also allows unsupervised learning of point-cloud based shapes by using a variational autoencoder, leading to higher-quality generated shapes.

## 1 Introduction

One of the challenges in 3D shape processing concerns the question of representation. Shapes are typically represented as triangle meshes or point clouds in computer graphics applications due to their simplicity and light-weight nature. At the same time an increasing number of robotic and remote-sensing applications are deploying sensors that directly collect point-cloud representations of the environment. Hence architectures that efficiently operate on point clouds are becoming increasingly desirable.

On the other hand the vast majority of computer vision techniques rely on grid-based representation of 3D shapes for analyzing and generating them. These include multiview representations that render a shape from a collection of views [31,39,37] or voxel-based representations [44,19,30,4,43] that discretize point occupancy onto a 3D grid. Such representations allow the use of convolution and pooling operations for efficient processing. However, voxel-representations scale poorly with resolution and are inefficient for modeling surface details. Even multiscale or sparse variants [25,33,17] incur relatively high processing cost. Image-based representations, while more efficient, are not effective at modeling shapes with concave or filled interiors due to self occlusions. Moreover, generating shapes as a collection of views requires subsequent reasoning about geometric consistency to infer the 3D shape, which can be challenging.

The main contribution of our work is a multiresolution tree network capable of both recognizing and generating 3D shapes directly as point clouds. An overview of the network and how it can be applied to different applications are shown in Figure 1. Our approach represents a 3D shape as a set of locality-preserving 1D ordered list of points at multiple resolution levels. We can obtain such a ordering by using space-partitioning trees such as kd-tree or rp-tree. Feed-forward processing on the underlying tree can be implemented as 1D convolutions and pooling on the list. However, as our experiments show, processing the list alone is not sufficient since the 1D ordering distorts the underlying 3D structure of the shape. To ameliorate this problem we employ a multi-grid network architecture [22] where the representation at a particular resolution influences feed-forward processing at adjacent resolutions. This preserves the locality of point in the underlying 3D shape, improves information flow across scales, enables the network to learn a coarse-to-fine representation, and results in faster convergence during training. Our network outperforms existing point-based networks [40,24,32] that operate on position ($xyz$) information of points. Specifically, it obtains **91.7%** accuracy on the ModelNet40 classification task, while remaining efficient. It also outperforms similar graph networks that do not maintain multiresolution representations.

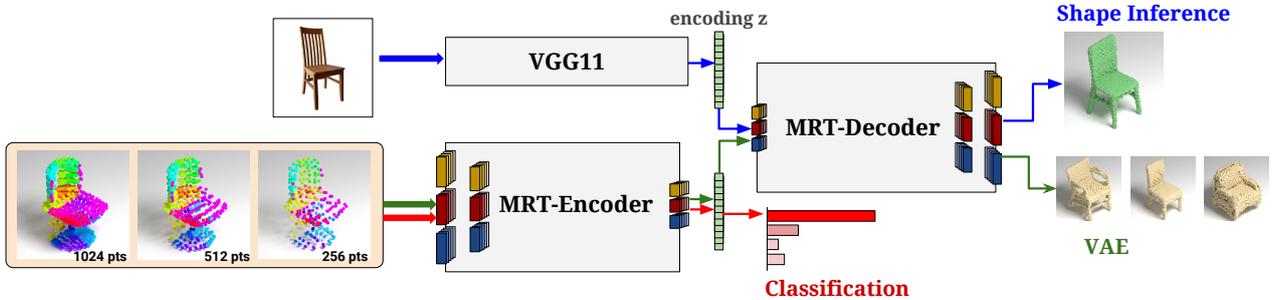

**Fig. 1:** Overview of MRTNet. On the left, the MRT-Encoder takes as input a 1D ordered list of points and represents it at multiple resolutions. Points are colored by their indices in the list. On the right, the MRT-Decoder directly outputs a point cloud. Our network can be used for several shape processing tasks, including classification (red), image-to-shape inference (blue), and unsupervised shape learning (green). Refer to Fig. 2 for details on the encoder and decoder.

Our multiresolution decoders can be used for directly generating point clouds. This allows us to incorporate order-invariant loss functions, such as Chamfer distance, over point clouds during training. Moreover it can can be plugged in with existing image-based encoders for image-to-shape inference tasks. Our method is able to both preserve the overall shape structure as well as fine details. On the task of single-image shape inference using the ShapeNet dataset, our approach outperforms existing voxel-based [9], view-based [26], and point-based [12] techniques.

Finally, the combined encoder-decoder network can be used for unsupervised learning of shape representations in a variational autoencoder (VAE) framework. The features extracted from the encoder of our VAE (trained on the unlabeled ShapeNet dataset) leads to better shape classification results (**86.4%** accuracy on ModelNet40) compared to those extracted from other unsupervised networks [43].

## 2   Related Work

A number of approaches have studied 3D shape recognition and generation using uniform 3D voxel grids [9,44,19,30,4,43]. However, uniform grids have poor scalability and require large memory footprint, hence existing networks built upon them often operate on a relatively low-resolution grid. Several recent works address this issue through multiscale and sparse representations [33,41,42,17,38,16] at the expense of additional book keeping. Still, voxel-based methods generally incur high processing cost, and are not well suited for modeling fine surface details. Moreover, it's not clear how to incorporate certain geometric attributes, like surface normals, into voxel representation, since these attributes do not exist in the interior of the shape.

Multiview methods [39,31,37,28,20] represent a 3D shape as images rendered from different viewpoints. These methods use efficient convolutional and pooling operations and leverage deep networks pretrained on large labeled image datasets. However, they are not optimal for general shapes with complex interior structures due to self occlusions. Nonetheless since most models on existing shape repositories are described well by their exterior surface, view-based approaches have been adapted for shape classification and segmentation tasks. Recently they have also been used for generation where a set of depth and normal maps from different viewpoints are inferred using image-based networks, and have been successfully used for image to shape generation tasks [28,26]. However such approaches requires subsequent processing to resolve view inconsistencies and outliers which is a challenging task.

Previous work has also studied extensions of ConvNets to mesh surfaces such as spectral CNNs [5,45], geodesic CNNs [29], or anisotropic CNNs [2]. They have shown success for local correspondence and matching tasks. However, some of these methods are constrained on manifold surfaces, and generally it's unclear how well they perform on shape generation tasks. A recent work in [35] generalized the convolution operator from regular grid to arbitrary graphs while avoiding the spectral domain, allowing graphs of varying size and connectivity.

Another branch of recent works focused on processing shapes represented as point clouds. One example is PointNet [40,32], that directly consumes point clouds as input. The main idea is to first process each point identically and independently, then leverage a symmetric function (max pooling) to aggregate information from all points. The use of max pooling preserves the permutation invariance of a point set, and the approach is quite effective for shape classification and segmentation tasks. Similarly, KD-net [24] operates directly on point cloud shapes. It spatially partitions a point cloud using a kd-tree, and imposes a feed-forward network on top of the tree structure. This approach is scalable, memory efficient, achieves competitive performance on shape recognition tasks. While successful as encoders, it hasn't been shown how these networks can be employed as decoders for shape generation tasks.

Generating shapes as a collection of points without intermediate modeling of view-based depth maps has been relatively unexplored in the literature. The difficulty stems from the lack of scalable approaches for generating sets. Two recent works are in this direction. Fan et al. [12] train a neural network to generate point clouds from a single image by minimizing Earth Mover's Distance (EMD) or Chamfer distance (CD) between the generated points and the model points. These distances are order invariant and hence can operate directly on point sets. This approach uses a two-branched decoder, one branch is built with 2D transposed convolutions and the other one is composed by fully connected layers. On the other hand, our approach uses a simpler and shallower decoder built as a composition of 1D deconvolutions that operate at multiple scales. This representation improves information flow across scales, which leads to higher quality generated shapes. Moreover, we use permutation invariant losses along with regularization of latent variables to build a model similar to a variational autoencoder [23] that can be used to sample shapes from gaussian noise. Another work in [13] learns a distribution over shape coefficients using a learned basis for a given category using a generative adversarial network [15]. However, in this approach, the underlying generative model assumes a linear shape basis, which produces less detailed surfaces. The improved scalability of our method allows generating shapes with more points and more accurate geometric details in comparison to previous work.

Our tree network builds on the ideas of multiscale [18,27], mutligrid [22] and dilated [46] or atrous filters [11,8] effective for a number of image recognition tasks. They allow larger receptive fields during convolutions with a modest increase in the number of parameters. In particular Ke et al. [22] showed that communication across multiresolutions of an image throughout the network leads to improved convergence and better accuracy on a variety of tasks. Our approach provides an efficient way of building multigrid-like representations for 3D point clouds.

## 3   Method

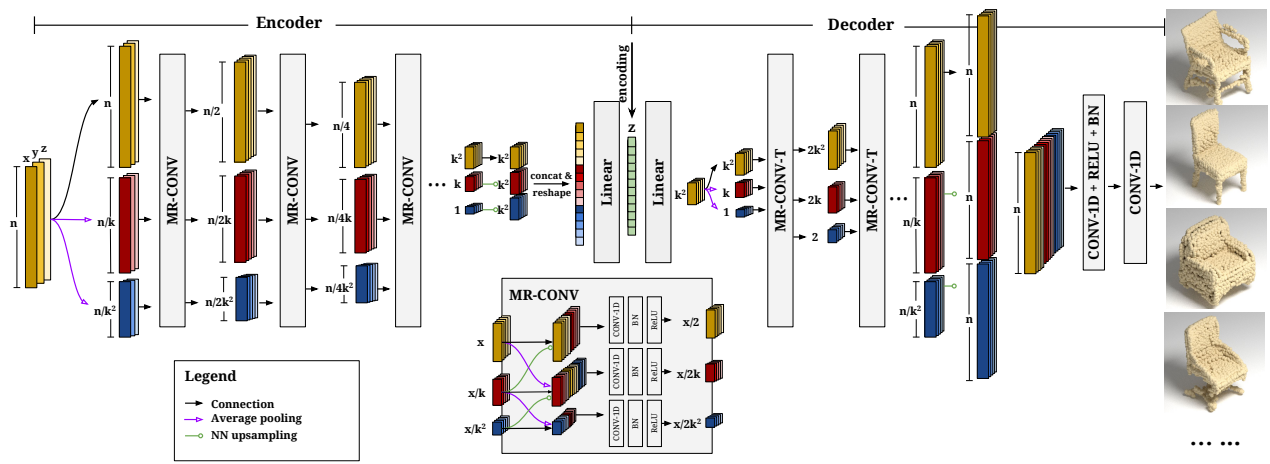

**Fig. 2:** Our multiresolution tree network (MRTNet) for processing 3D point clouds. We represent a 3D shape as a 1D list of spatially sorted point cloud. The network represents each layer at three scales (indicated by yellow, red, and blue), the scale ratio is $k$ between each two. The last two convolution layers have kernel size 1 and stride 1. MR-CONV refers to multi-resolution convolution (zoom-in to the inset for details); and MR-CONV-T refers to MR-CONV with transposed convolution. Our network is flexible and can be used for several shape processing tasks.

Figure 2 shows the complete architecture of our multiresolution tree network (MRTNet) that includes both the encoder and decoder. We represent 3D shapes as a point cloud of a fixed size $N = 2^D$ (e.g. $N = 1K$). We center the point cloud at the origin and normalize its bounding box; then spatially sort it using a space-partitioning tree. The input to the network are thus a 1D list ($N \times 3$ tensor) containing the $xyz$ coordinates of the points. The network leverages 1D convolution and represents each layer at three scales, with a ratio of $k$ between each two. MR-CONV refers to multi-resolution convolution, and MR-CONV-T refers to MR-CONV with transposed convolution. The encoding **z** is a 512-D vector. Our network architecture is flexible and can be used for several shape processing tasks. For shape **classification**, we use only the multiresolution encoder but adding a fully connected layer after the encoding **z** to output a 40-D vector representing the ModelNet40 shape classes. For **single-image shape inference**, we employ a pretrained VGG-11 image encoder [36], combined with our multiresolution decoder to directly an output point cloud shape as a $N \times 3$ tensor. For **unsupervised learning of point clouds**, we use both the multiresolution encoder and decoder, forming a variational autoencoder.

**Spatial sorting.** As a point cloud is unordered to begin with, we use a space-partitioning tree such as KD-tree to order the points. To start, we sort the entire point set along the $x$-axis, then split it in half, resulting in equal-sized left and right subsets; we then recursively split each subset, this time along the $y$-axis; then along $z$-axis; then back along the $x$-axis and so on. Basically it's a recursive process to build a full tree where the splitting axes alternate between $x$, $y$, $z$ at each level of the tree. The order of leaf nodes naturally becomes the order of the points. There are several variants on the splitting strategy. If at each split we choose an axis among $x$, $y$, $z$ with probability proportional to the span of the subset along that axis, it builds a probabilistic KD-tree as described in [24]. If we choose axes from a random set of directions, it builds an RP-tree [10]. Note that after the ordering is obtained, the underlying details of the how the splits were taken are discarded. This is fundamentally different from [24] where the network computations are conditioned on the splitting directions.

The purpose of spatial sorting is to build a hierarchical and locality-preserving order of the points. Thus functions computed based on the local 3D neighborhood at a point can be approximated using convolutions and pooling operations on the 1D structure. However, any ordering of points is distortion inducing and in particular long-range relationships are not preserved well. Maintaining multiple resolutions of the data allows us to preserve locality at different scales. Since the partitioning is constructed hierarchically this can be efficiently implemented using pooling operations described next.

**Multiresolution convolution.** With the spatially sorted point set, we can build a network using 1D convolution and pooling operations. The convolution leverages the spatial locality of points after sorting, and the pooling leverages the intrinsic binary tree structure of the points.

With a conventional CNN, each convolutional operation has a restricted receptive field and is not able to leverage both global and local information effectively. We resolve this issue by maintaining three different resolutions of the same point cloud using a mutligrid architecture [22]. Different resolutions are computed directly through pooling and upsampling operations. Specifically, we use average pooling with kernel size and stride of $k$, where $k$ is a power of 2. This configuration allows pooling/downsampling the point cloud while preserving its hierarchical tree structure. Figure 1 (left) shows an example point cloud at three different resolutions computed by pooling with $k = 2$. For upsampling, we use nearest neighbor (NN) upsampling with a factor of $k$.

Once we can pool and upsample the point clouds, we are able to combine global and local information in the convolutional operations by using the MR-CONV block in the inset of Fig. 2. The multiresolution block operates in the following way. We maintain the point cloud representations at three resolutions $\mathbf{f}_{(0)}, \mathbf{f}_{(1)}, \mathbf{f}_{(2)}$, where the scale ratio between each two is (as mentioned above) $k$. The MR-CONV block receives all three as input, and each resolution will be upsampled and/or pooled and concatenated with each other, creating three new representations $\mathbf{f}'_{(0)}, \mathbf{f}'_{(1)}, \mathbf{f}'_{(2)}$:

$$\mathbf{f}'_{(0)} = \mathbf{f}_{(0)} \oplus up(\mathbf{f}_{(1)}); \quad \mathbf{f}'_{(1)} = pool(\mathbf{f}_{(0)}) \oplus \mathbf{f}_{(1)} \oplus up(\mathbf{f}_{(2)}); \quad \mathbf{f}'_{(2)} = pool(\mathbf{f}_{(1)}) \oplus \mathbf{f}_{(2)}.$$

where $\oplus$ is the concatenation operation, $up$ and $pool$ are the upsampling and average pooling operations. Each new representation $\mathbf{f}'$ then goes through a sequence of operations: 1D convolution (kernel size=2 and stride=2), batch normalization and ReLU activation. Note that due to the stride 2, each output is half the size of its associated input. In our generative model and shape inference model we use $k = 4$, while for classification we use $k = 8$.

**Shape classification model.** For classification, we use our multiresolution encoder in Figure 2, and add a fully connected layer after encoding $\mathbf{z}$ that outputs a 40-D vector representing the ModelNet40 classification. Specifically, we train the network on the ModelNet40 [44] dataset, which contains 12,311 objects covering 40 different categories. It is split into 9,843 shapes for training and 2,468 shapes for testing. For each object, we sample 1K points on the surface using Poisson Disk sampling [3] to evenly disperse the points. Each sampled point cloud is then spatially sorted using the probabilistic kd-tree [24]. Specifically, at each split of the tree we choose a random split axis according to the following PDF:

$$P(\mathbf{n} = \mathbf{e}_i | \mathbf{x}) = \frac{\exp\{span_i(\mathbf{x})\}}{\sum_{j=1}^{d} \exp\{span_j(\mathbf{x})\}}$$

where $\mathbf{x}$ is the subset of points to be split, $\mathbf{n}$ is the split axis chosen from the canonical axes $\mathbf{e}_i$ (i.e. $x,y,z$ in 3D), and $span_i(\mathbf{x})$ returns the span of $\mathbf{x}$ along each axis $\mathbf{e}_i$.

The network parameters are as follows: the first MR-CONV layer has 16 filters and the following layers double the amount of filter of the previous one, unless the previous layer has 1024 filters. In that case, the next layer also has 1024 filters. The network is trained by minimizing a cross-entropy loss using an Adam optimizer with learning rate $10^{-3}$ and $\beta = 0.9$. The learning rate is decayed by dividing it by 2 every 5 training epochs. We employ scale augmentation at training and test time by applying anisotropic scaling factors drawn from $\mathcal{N}(1, 0.25)$. At test time, for each point cloud we apply the sampled scale factors and build the probabilistic kd-tree 16 times as described above, thus obtaining 16 different versions and orderings of the point set. Our final classification is the mean output of those versions. The test-time average has very little impact on the computation time (a discussion is included in Sec. 4.4).

**Single-image shape inference.** Our multiresolution decoder can be used to perform image-to-shape inference. Specifically, we use a pretrained VGG-11 image encoder [36] combined with our decoder in Figure 2. Our decoder is set to generates 4K points. The entire network is trained using the dataset and splits provided by [9], which contains 24 renderings from different views for 43783 shapes from ShapeNet divided in 13 different categories. We sample each ShapeNet mesh at 4K points and use them for supervision. Given a rendered image, the task is to predict the complete point cloud (4K points) representing the object in the image. The decoder in Figure 2 has the following number of filters per layer: 512-512-256-256-128-64-64-64. As in Figure 2, the two additional convolutional layers at the end have kernel size 1 and stride 1: the first one has 128 filters and the second one outputs the final 4K point set.

There are many possible choices for the reconstruction loss function. One straightforward choice would be to use the ordering induced by the spatial partitioning and compute the $L_2$ loss between the output and ground-truth point clouds. However, $L_2$ loss turns out to work very poorly. We chose to use the Chamfer distance between two point clouds ($\mathbf{x}$ and $\mathbf{y}$), defined as:

$$Ch(\mathbf{x}, \mathbf{y}) = \frac{1}{|\mathbf{x}|} \sum_{x \in \mathbf{x}} \min_{y \in \mathbf{y}} \|x - y\|_2 + \frac{1}{|\mathbf{y}|} \sum_{y \in \mathbf{y}} \min_{x \in \mathbf{x}} \|x - y\|_2$$

The Chamfer distance is invariant to permutations of points, making it suitable to measure dissimilarities between unstructured point clouds. The model is trained using an Adam optimizer with learning rate $10^{-3}$ and $\beta = 0.9$. Learning rate is divided by two at each two epochs.

**Unsupervised learning of point clouds.** By combining the multiresolution encoder and decoder together, we can perform unsupervised learning of 3D shapes. The entire network, called MR-VAE, builds upon a variational autoencoder (VAE) [23] framework. The encoder $Q$ receives as an input a point cloud $\mathbf{x}$ and outputs an encoding $\mathbf{z} \in \mathbb{R}^{512}$. The decoder $D$ tries to replicate the point cloud $\mathbf{x}$ from $\mathbf{z}$. Both encoder and decoder are built using a sequence of MR-CONV blocks as in Fig. 2. Similar to above, we use Chamfer distance as the reconstruction loss function. Besides this, we also need a regularization term that forces the distribution of the encoding $\mathbf{z}$ to be as close as possible to the Gaussian $\mathcal{N}(0, I)$. Differently from the original VAE model, we found that we can get more stable training if we try to match the first two moments (mean and variance) of $\mathbf{z}$ to $\mathcal{N}(0, I)$. Mathematically, this regularization term is defined as:

$$\mathcal{L}_{reg} = \|cov(Q(\mathbf{x}) + \delta) - I\|_2 + E[Q(\mathbf{x}) + \delta]$$

where $cov$ is the covariance matrix, $Q$ is the encoder, $\|\cdot\|_2$ is the Frobenius norm and $E[\cdot]$ is the expected value. $\delta$ is a random value sampled from $\mathcal{N}(0, cI)$ and $c = 0.01$. Thus, our generative model is trained by minimizing the following loss function:

$$\mathcal{L} = Ch(\mathbf{x}, D(Q(\mathbf{x}))) + \lambda L_{reg}$$

We set $\lambda = 0.1$. The model is trained using an Adam optimizer with learning rate $10^{-4}$ and $\beta = 0.9$. The encoder follows the classification model and the decoder follows the one used in the shape inference model, both described previously.

**Shape part segmentation.** MRTNet can also be applied for shape part segmentation tasks. For details please refer to the supplemental material.

## 4 Experimental Results and Discussions

This section presents experimental results. We implemented MRTNet using PyTorch.

| Method | Accuracy |
|---|---|
| *View-based methods* | |
| MVCNN [39] | 90.1 |
| MVCNN-MultiRes [31] | 91.4 |
| *Point-based methods (w/o normals)* | |
| KDNet (1K pts) [24] | 90.6 |
| PointNet (1K pts) [40] | 89.2 |
| PointNet++ (1K pts) [32] | 90.7 |
| MRTNet (1K pts) | **91.2** |
| MRTNet (4K pts) | **91.7** |
| KDNet (32K pts) [24] | **91.8** |
| *Point-based methods (with normals)* | |
| PointNet++ (5K pts) [32] | 91.9 |
| *Voxel-based methods* | |
| OctNet [33] | 86.5 |
| O-CNN [42] | 90.6 |

(a) **Comparisons with previous work**. Among point-based methods that use $xyz$ data only, ours is the best in the 1K points group; and our 4K result is comparable with KDNet at 32K points.

| Method | Accuracy |
|---|---|
| Full model (MRTNet, 4K pts) | 91.7 |
| Filters/4 | 91.7 |
| Single res. | 89.3 |
| Single res., no aug. (kd-tree) | 86.2 |
| Single res., no aug. (rp-tree) | 87.4 |

(b) **MRTNet ablation studies**. Filters/4 reduces the number of filters in each layer by 4. The last three rows are the single resolution model.

| Method | Accuracy |
|---|---|
| SPH [21] | 68.2 |
| LFD [7] | 75.5 |
| T-L Network [14] | 74.4 |
| VConv-DAE [34] | 75.5 |
| 3D-GAN [43] | 83.3 |
| MRTNet-VAE (Ours) | **86.4** |

(c) **Unsupervised representation learning**. Section 4.3.

Table 1: Instance classification accuracy on the ModelNet40 dataset.

## 4.1 Shape classification

To demonstrate the effectiveness of the multiresolution encoder, we trained a baseline model that follows the same classification model but replacing multiresolution convolutions with single-scale 1D convolutions. Also, we apply the same test-time data augmentation and compute the test-time average as described in the Section 3.

Classification benchmark results are in Table 1(a). As shown in the table, MRTNet achieves the best results among all **point-based** methods that use $xyz$ data only. In particular, ours is the best in the 1K points group. We also experimented with sampling shapes using 4K points, and the result is comparable with KDNet at 32K points – in this case, KDNet uses 8× more points (hence 8× more memory) than ours, and is only 0.1% better. PointNet++ [32] with 5K points and normals is 0.2% better than ours.

Table 1(b) shows ablation study results with variants of our approach. Particularly, the multiresolution version is more than 2% better than the baseline model (i.e. single resolution), while using the same number of parameters (the Filters/4 version). Besides, MRTNet converges must faster than the baseline model, as we can see in the cross entropy loss decay plots in Figure 3. This shows that the multiresolution architecture leads to higher quality/accuracy and is memory efficient.

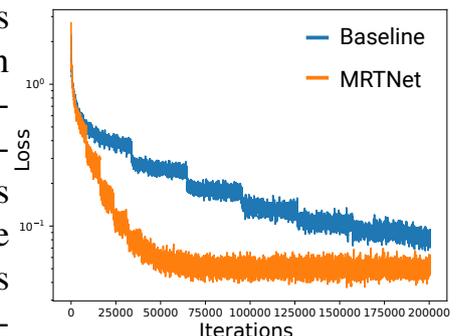

Fig. 3: Cross entropy decay

| Category | 3D-R2N2 [9] | | | Fan et al. [12] (1 view) | Lin et al. [26] (1 view) | MRTNet (1 view) |
|---|---|---|---|---|---|---|
| | 1 view | 3 views | 5 views | | | |
| airplane | 3.207 / 2.879 | 2.521 / 2.468 | 2.399 / 2.391 | 1.301 / 1.488 | 1.294 / 1.541 | **0.976 / 0.920** |
| bench | 3.350 / 3.697 | 2.465 / 2.746 | 2.323 / 2.603 | 1.814 / 1.983 | 1.757 / 1.487 | **1.438 / 1.326** |
| cabinet | 1.636 / 2.817 | 1.445 / 2.626 | **1.420** / 2.619 | 2.463 / 2.444 | 1.814 / **1.072** | 1.774 / 1.602 |
| car | 1.808 / 3.238 | 1.685 / 3.151 | 1.664 / 3.146 | 1.800 / 2.053 | 1.446 / **1.061** | **1.395** / 1.303 |
| chair | 2.759 / 4.207 | 1.960 / 3.238 | 1.854 / 3.080 | 1.887 / 2.355 | 1.886 / 2.041 | **1.650 / 1.603** |
| display | 3.235 / 4.283 | 2.262 / 3.151 | 2.088 / 2.953 | 1.919 / 2.334 | 2.142 / **1.440** | **1.815** / 1.901 |
| lamp | 8.400 / 9.722 | 6.001 / 7.755 | 5.698 / 7.331 | 2.347 / 2.212 | 2.635 / 4.459 | **1.944 / 2.089** |
| speaker | 2.652 / 4.335 | 2.577 / 4.302 | 2.487 / 4.203 | 3.215 / 2.788 | 2.371 / **1.706** | **2.165** / 2.121 |
| rifle | 4.798 / 2.996 | 4.307 / 2.546 | 4.193 / 2.447 | 1.316 / 1.358 | 1.289 / 1.510 | **1.029 / 1.028** |
| sofa | 2.725 / 3.628 | 2.371 / 3.252 | 2.306 / 3.196 | 2.592 / 2.784 | 1.917 / **1.423** | **1.768** / 1.756 |
| table | 3.118 / 4.208 | 2.268 / 3.277 | 2.128 / 3.134 | 1.874 / 2.229 | 1.689 / 1.620 | **1.570 / 1.405** |
| telephone | 2.202 / 3.314 | 1.969 / 2.834 | 1.874 / 2.734 | 1.516 / 1.989 | 1.939 / **1.198** | **1.346** / 1.332 |
| watercraft | 3.592 / 4.007 | 3.299 / 3.698 | 3.210 / 3.614 | 1.715 / 1.877 | 1.813 / 1.550 | **1.394 / 1.490** |
| **mean** | 3.345 / 4.102 | 2.702 / 3.465 | 2.588 / 3.342 | 1.982 / 2.146 | 1.846 / 1.701 | **1.559 / 1.529** |

**Table 2: Single-image shape inference results.** The training data consists of 13 categories of shapes provided by [9]. The numbers shown are [pred→GT / GT→pred] errors, scaled by 100. The mean is computed across all 13 categories. Our MRTNet produces 4K points for each shape.

| Fully Connected | Single Res. | MRTNet |
|---|---|---|
| 1.824 / 2.297 | 1.708 / 1.831 | **1.559 / 1.529** |

**Table 3: Ablation studies for the image to shape decoder.** The numbers shown are [pred→GT / GT→pred] errors, scaled by 100. The values are the mean computed across all 13 categories.

Our single resolution baseline is akin to KDNet except it doesn't condition the convolutions on the splitting axes. It results in 1.3% less classification accuracy compared to KDNet (1K pts). This suggests that conditioning on the splitting axes during convolutions improves the accuracy. However, this comes at the cost of extra book keeping and at least three times more parameters. MRTNet achieves greater benefits with lesser overhead. Similar to the KDNet, our methods also benefit from data augmentation and can be used with both kd-trees and rp-trees.

### 4.2 Single-image shape inference

We compare our single-image shape inference results with volumetric [9], view-based [26] and point-based [12] approaches using the evaluation metric by [26]. Given a source point cloud **x** and a target point cloud **y**, we compute the average euclidean distance from each point in **x** to its closest in **y**. We refer to this as pred→GT (prediction to groundtruth) error. It indicates how dissimilar the predicted shape is from the groundtruth. The GT→pred error is computed similarly by swapping **x** and **y**, and it measures coverage (i.e. how complete the ground-truth surface was covered by the prediction). For the voxel based model [9], we used the same procedure as [26], where point clouds are formed by creating one point in the center of each surface voxel. Surface voxels are extracted by subtracting the prediction by its eroded version and rescale them such that the tightest 3D bounding boxes of the prediction and the ground-truth CAD models have the same volume.

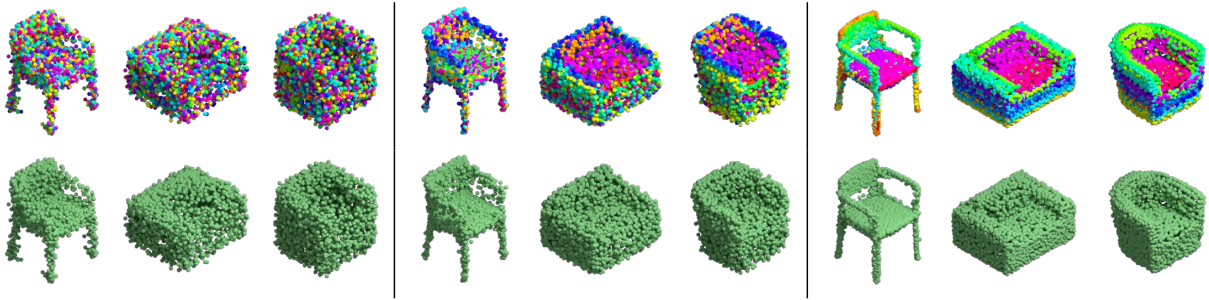

**Fig. 4:** Shapes generated by 1) the fully connected baseline; 2) the single-resolution baseline; and 3) MRTNet. Colors in the first row indicate the index of a point in the output point list.

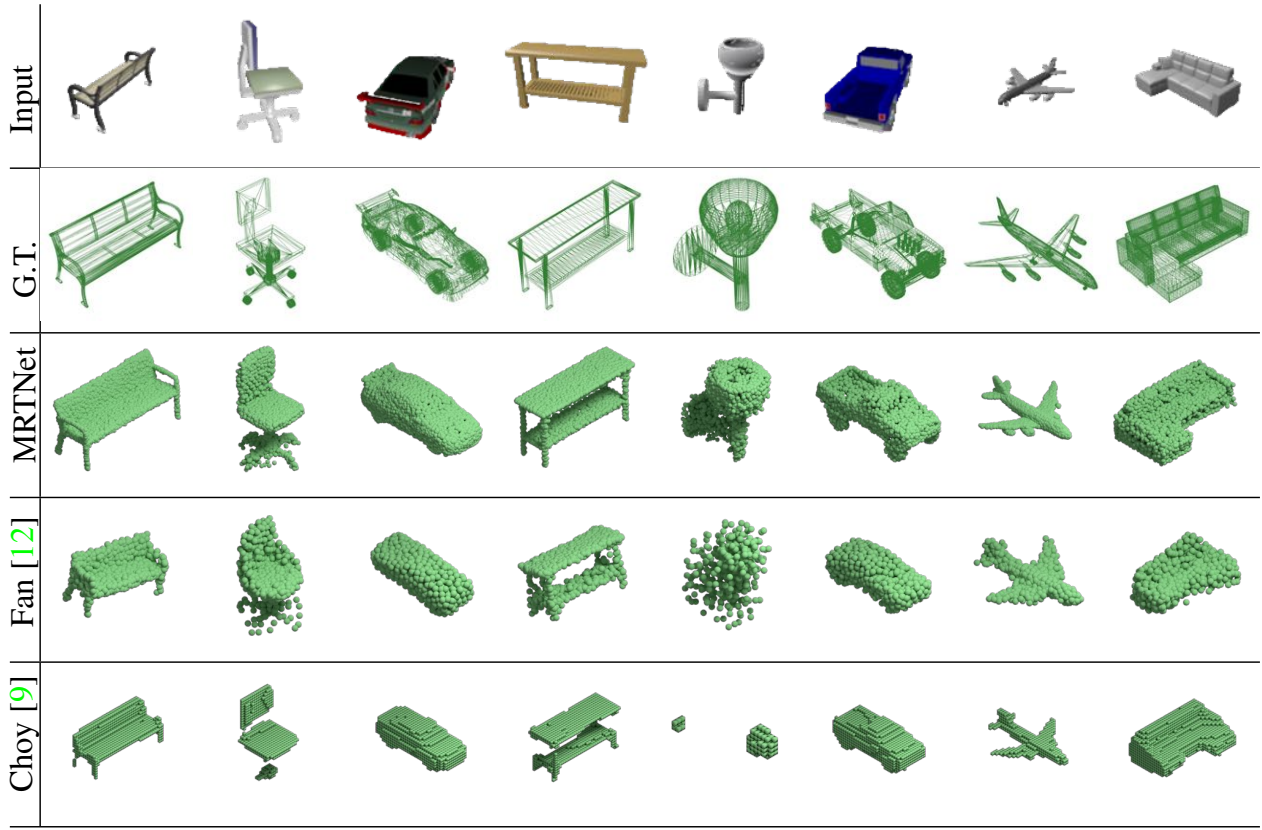

**Fig. 5:** Qualitative results for single-image shape inference. From top to bottom: input images, ground truth 3D shapes, results of MRTNet, Fan et al. [12], and Choy et al. [9].

Table 2 shows our results. Our solution outperforms competing methods in 12 out of 13 categories on the pred→GT error, and in 6 categories on GT→pred error. Note that we are consistently better than the point-based methods such as [12] in both metrics; and we are consistently better than [26] in the pred→GT metric. Furthermore, our method wins by a considerable margin in terms of the mean per category on both metrics. It is important to highlight that the multi-view based method [26] produces tens of thousands of points and many of them are not in the right positions, which penalizes their pred→GT metric, but that helps to improve their GT→pred. Moreover, as mentioned in [26], their method has difficulties capturing thin structures (e.g. lamps) whereas ours is able to capture them relatively well. For example, our GT→pred error for the **lamp** category (which contains many thin geometric structures) is more than two times smaller than the error by [26], indicating that MRTNet is more successful at capturing thin structures in the shapes.

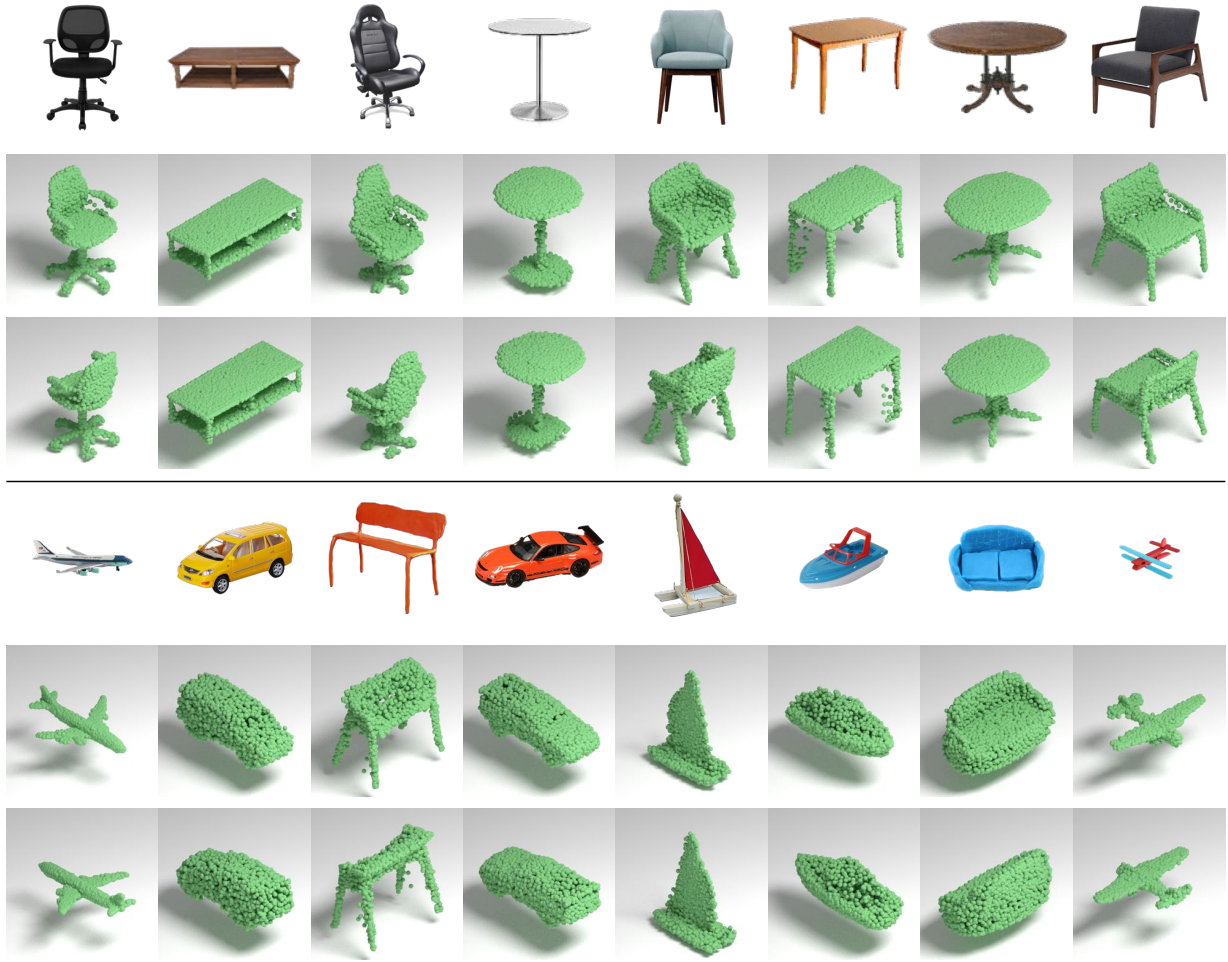

**Fig. 6:** Shapes generated by applying MRTNet on Inernet photos of furnitures and toys. MRTNet is trained on the 13 categories of ShapeNet database (Table 2) . Note how the network is capable of generating detailed shapes from real photos, even though it is trained only on rendered images using simple shading models. For each output shape we show two different views.

**Ablation studies.** In order to quantify the effectiveness of the multiresolution decoder, we compared our method with two different baselines: a fully connected decoder and a single-resolution decoder. The fully connected decoder consists of 3 linear layers with 4096 hidden neurons, each layer followed by batch normalization and ReLU activation units. On top of that, we add a final layer that outputs $4096 \times 3$ values corresponding to the final point cloud, followed by a hyperbolic tangent activation function. The single resolution decoder follows the same architecture of the MRT decoder but replacing multiresolution convolutions with single-scale 1D convolutions. Results are shown in Table 3. Note that both baselines are quite competitive. The single-resolution decoder is comparable to the result of [26], while the fully connected one achieves similar mean errors to [12]. Still, they fall noticeably behind MRTNet.

In Figure 4 we visualize the structures of the output point clouds generated by the three methods. The point clouds generated by MRTNet present strong spatial coherence: points that are spatially nearby in 3D are also likely to be nearby in the 1D list. This coherence is present to some degree in the single-resolution outputs (note the dark blue points in the chair's arms), but is almost completely absent in the results by the fully connected decoder. This is expected, since fully connected layers do not leverage the spatial correlation of their inputs. Operating at multiple scales enables MRTNet to enforce a stronger spatial coherence, allowing it to more efficiently synthesize detailed point clouds with coherent geometric structures.

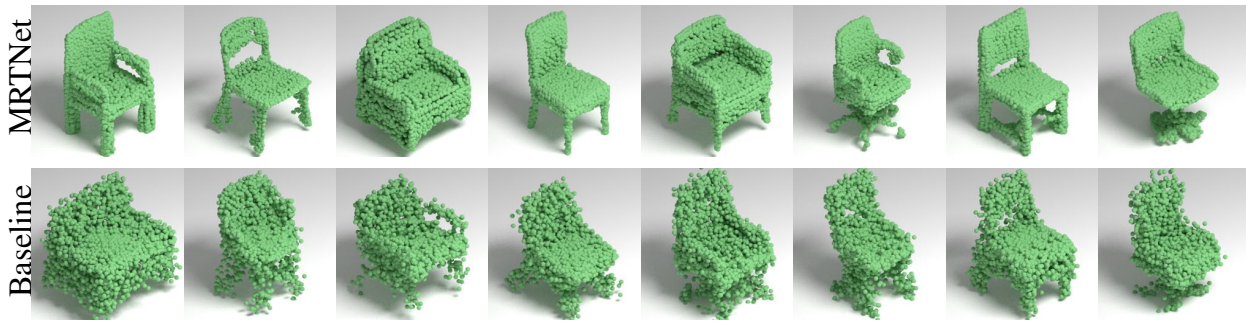

**Fig. 7:** Qualitative comparisons of MR-VAE with a single-resolution baseline model. Results are generated by randomly sampling the encoding **z**. MR-VAE is able to preserve shape details much better than the baseline model, and produces less noisy outputs.

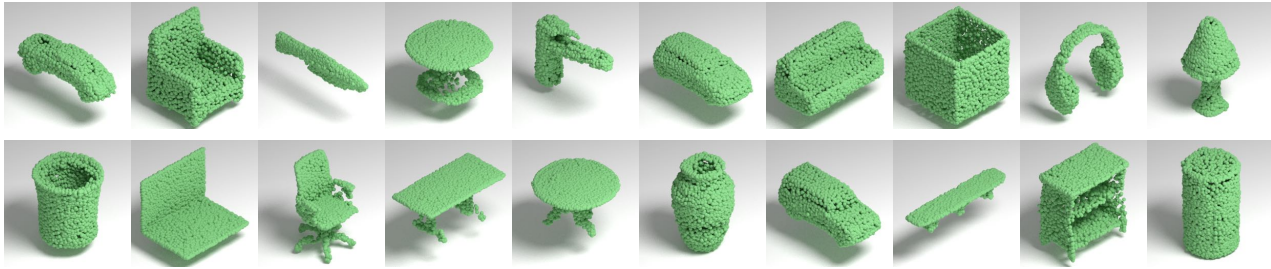

**Fig. 8:** Test set shapes reconstructed by MR-VAE trained on all categories of ShapeNet (using 80%/20% training/test split). MR-VAE is able to reconstruct high-quality diverse shapes.

**Qualitative results.** In Figure 5 we present qualitative results of our method and comparisons to two prior works. The input images have 3 color channels and dimensions $224 \times 224$. In Figure 6 we show results of our method applied on photographs downloaded from the Internet. To apply our method, we manually removed the background from the photos using [1], which generally took less than half a minute per photo. As seen from the results, MRTNet is able to capture the structure and interesting geometric details of the objects (e.g. wheels of the office chairs), even though the input images are considerably different from the rendered ones used in training.

### 4.3   Unsupervised Learning of Point Clouds

For unsupervised learning of point clouds, we train our MR-VAE using the ShapeNet dataset [6]. By default we compute $N = 4K$ points for each shape using Poisson Disk sampling [3] to evenly disperse the points. Each point set is then spatially sorted using a kd-tree. Here we use the vanilla kd-tree where the splitting axes alternate between $x$, $y$, $z$ at each level of the tree. The spatially sorted points are used as input to train the MR-VAE network (Section 3). Similar to before, we also train a baseline model that follows the same network but replacing multiresolution convolutions with single-scale 1D convolutions in both encoder and decoder. As Figure 7 shows, the shapes generated by the MR-VAE trained on chairs are of considerably higher quality than those generated by the baseline model.

We also performed multiple-category shape generation by training MR-VAE on 80% of the objects from ShapeNet dataset. The remaining models belong to our test split. Reconstructions of objects in the test split are included in Figure 8. Even when trained with a greater variety of shapes, the MR-VAE is able to reconstruct high quality shapes from its embedding. This demonstrates that MR-VAE is suitable for various inference tasks such as shape completion or point cloud reconstructions.

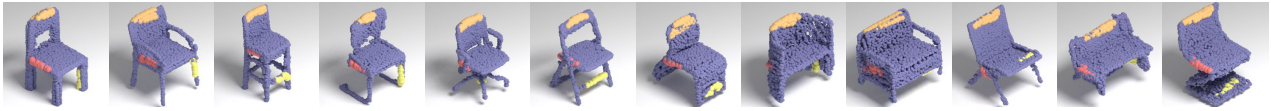

**Fig. 9:** Point correspondences among different shapes generated by MR-VAE. We picked three index ranges (indicated by three colors) from one example chair, and then color coded points in every shape that fall into these three ranges. The images clearly show that the network learned to generate shapes with consistent point ordering.

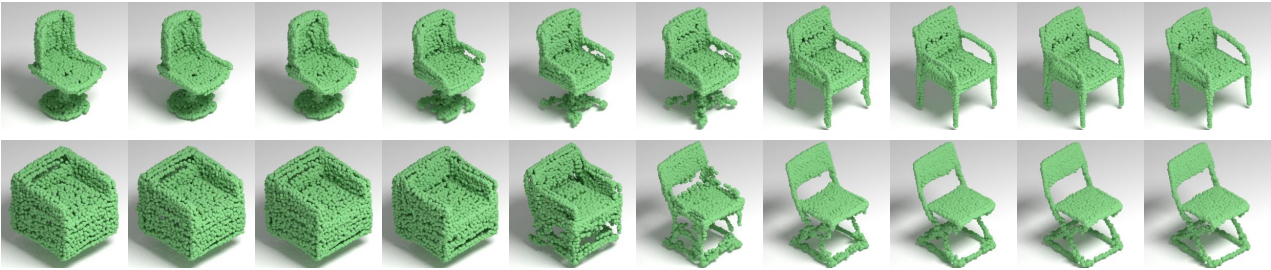

**Fig. 10:** Shape interpolation results. For each example, we obtain the encodings **z** of the starting shape and ending shape, then linearly interpolate the encodings and use the decoder to generate output shapes from the interpolated **z**. Results show plausible interpolated shapes.

**Point ordering in the generated shapes.** A useful way to analyze shape generation is to see if the generated points have any consistent ordering across different shapes. This is an interesting question because as described previously, our MR-VAE is trained using Chamfer Distance, a metric that's invariant to permutations of points. While the input to the network is all spatially sorted, the output is not restricted to any particular order and can in theory assume any arbitrary order. In practice, similar to the image-to-shape model, we observe that there is a consistent ordering of points in the generated shapes, as shown in Figure 9. Specifically, we picked three index ranges from one example chair, one at the top, one on the side, and one close to the bottom, then we color coded points in each shape that fall into these three index ranges. In the figure we can see clearly that they fall into approximately corresponding regions on each chair shape.

**Shape interpolation.** Another common test is shape interpolation: pick two encodings (either randomly sampled, or generated by the encoder for two input shapes), linearly interpolate them and use the decoder to generate the output shape. Figure 10 shows two sets of interpolation results of chairs from the ShapeNet dataset.

**Unsupervised classification.** A typical way of assessing the quality of representations learned in a unsupervised setting is to use them as features for classification. To do so, we take the MR-VAE model trained with all ShapeNet objects, and use its features to classify ModelNet40 [44] objects. Our classifier is a single linear layer, where the input is a set of features gathered from the first three layers of the MR-VAE encoder. The features are constructed this way: we apply a pooling operation of size 128, 64 and 32 respectively on these three layers; then at each layer upsample the two smaller resolutions of features to the higher resolution such that all three resolutions have the same size. Finally, we concatenate all those features and pass them through a linear layer to get the final classification. It is important to notice that we did not perform any fine-tuning: the only learned parameters are those from the single linear layer. We used an Adam optimizer with learning rate $10^{-3}$ and $\beta = 0.9$. The learning rate is decayed by dividing it by 2 every 5 epochs. Using this approach, we obtained an accuracy of $86.34\%$ on the ModelNet40 classification benchmark, as shown in Table 1(c). This result is considerably higher compared to similar features extracted from unsupervised learning in other autoencoders. This shows that the representations learned by our MR-VAE is more effective at capturing and linearizing the latent space of training shapes.

### 4.4 Discussions

**Robustness to transformations.** Kd-trees are naturally invariant to point jittering as long as it's small enough so as to not alter the shape topology. Our approach is invariant to translations and uniform scaling as the models are re-centered at the origin and resized to fit in the unit cube. On the other hand, kd-trees are not invariant to rotations. This can be mitigated by using practices like pooling over different rotations (e.g. MVCNN) or branches that perform pose prediction and transformation (e.g. PointNet). However, we notice that simply having unaligned training data was enough to account for rotations in the classification task, and the ModelNet40 dataset contains plenty of unaligned shapes. Moreover, since the KDNet [24] also employs a kd-tree spatial data structure, the discussions there about transformations also apply to our method.

**Computation time.** Building a kd-tree of $N$ points takes $O(N \log N)$ time, where $N = 2^{10}$ for 1K points. While PointNet does not require this step, it's also more than 2.0% worse in the classification task. The time to run a forward pass for classification is as follows: PointNet takes 25.3ms, while MRTNet takes 8.0ms on a TITAN GTX1080, both with batch size of 8. Kd-tree building is also much faster than rendering a shape multiple times like in MVCNN [39] or voxelizing it [33]. Using 16 different test-time augmentations does not have significant impact in computational time, as the 16 versions are classified in the same batch. This number of test-time augmentations is comparable to other approaches, e.g. 10 in [24], 80 in [39], and 12 in [42] and [40].

## 5   Conclusion

In conclusion, we introduced multiresolution tree networks (MRTNet) for point cloud processing. They are flexible and can be used for shape classification, generation, and inference tasks. Our key idea is to represent a shape as a set of locality-preserving 1D ordered list of points at multiple resolutions, allowing efficient 1D convolution and pooling operations. The representation improves information flow across scales, enabling the network to perform coarse-to-fine analysis, leading to faster convergence during training and higher quality for shape generation.

In future work, we would like to incorporate additional point attributes, such as normal and color, into the network, to further improve accuracy of shape recognition and allow the shape generator to produce these attributes. We would also like to apply and extend the method to process spatio-temporal shape analysis, such as on animated shapes or other temporarlly changing 3D point data.

# References

1. <https://clippingmagic.com/>.
2. D. Boscaini, J. Masci, E. Rodolà, and M. M. Bronstein. Learning shape correspondence with anisotropic convolutional neural networks. In *NIPS*, 2016.
3. J. Bowers, R. Wang, L.-Y. Wei, and D. Maletz. Parallel poisson disk sampling with spectrum analysis on surfaces. *ACM Trans. Graph.*, 29(6):166:1–166:10, 2010.
4. A. Brock, T. Lim, J. M. Ritchie, and N. Weston. Generative and discriminative voxel modeling with convolutional neural networks. arXiv preprint arXiv:1608.04236, 2016.
5. J. Bruna, W. Zaremba, A. Szlam, and Y. LeCun. Spectral networks and locally connected networks on graphs. arXiv preprint arXiv:1312.6203, 2013.
6. A. X. Chang, T. Funkhouser, L. Guibas, P. Hanrahan, Q. Huang, Z. Li, S. Savarese, M. Savva, S. Song, H. Su, et al. Shapenet: An information-rich 3d model repository. *arXiv preprint arXiv:1512.03012*, 2015.
7. D.-Y. Chen, X.-P. Tian, Y.-T. Shen, and M. Ouhyoung. On Visual Similarity Based 3D Model Retrieval. *Computer Graphics Forum*, 2003.
8. L.-C. Chen, G. Papandreou, I. Kokkinos, K. Murphy, and A. L. Yuille. Deeplab: Semantic image segmentation with deep convolutional nets, atrous convolution, and fully connected crfs. *arXiv preprint arXiv:1606.00915*, 2016.
9. C. B. Choy, D. Xu, J. Gwak, K. Chen, and S. Savarese. 3D-R2N2: A unified approach for single and multi-view 3D object reconstruction. In *European Conference on Computer Vision*, 2016.
10. S. Dasgupta and Y. Freund. Random projection trees and low dimensional manifolds. In *Proceedings of the Fortieth Annual ACM Symposium on Theory of Computing*, pages 537–546, 2008.
11. P. Dutilleux. An implementation of the "algorithme à trous" to compute the wavelet transform. In *Wavelets*, pages 298–304. Springer, 1990.
12. H. Fan, H. Su, and L. Guibas. A point set generation network for 3d object reconstruction from a single image. In *IEEE Conference on Computer Vision and Pattern Recognition*, 2017.
13. M. Gadhela, S. Maji, and R. Wang. 3d shape generation using spatially ordered point clouds. In *British Machine Vision Conference (BMVC)*, 2017.
14. R. Girdhar, D. Fouhey, M. Rodriguez, and A. Gupta. Learning a predictable and generative vector representation for objects. In *ECCV*, 2016.
15. I. Goodfellow, J. Pouget-Abadie, M. Mirza, B. Xu, D. Warde-Farley, S. Ozair, A. Courville, and Y. Bengio. Generative adversarial nets. In *Advances in Neural Information Processing Systems (NIPS)*, 2014.
16. B. Graham and L. van der Maaten. Submanifold sparse convolutional networks. *arXiv preprint arXiv:1706.01307*, 2017.
17. C. Häne, S. Tulsiani, and J. Malik. Hierarchical surface prediction for 3d object reconstruction. In *International Conference on 3D Vision (3DV)*, 2017.
18. K. He, X. Zhang, S. Ren, and J. Sun. Spatial pyramid pooling in deep convolutional networks for visual recognition. In *European Conference on Computer Vision*, pages 346–361. Springer, 2014.
19. J. Huang and S. You. Point cloud labeling using 3d convolutional neural network. In *ICPR*, pages 2670–2675, 2016.
20. E. Kalogerakis, M. Averkiou, S. Maji, and S. Chaudhuri. 3D shape segmentation with projective convolutional networks. In *CVPR*, 2017.
21. M. Kazhdan, T. Funkhouser, and S. Rusinkiewicz. Rotation invariant spherical harmonic representation of 3d shape descriptors. In *Proceedings of the 2003 Eurographics/ACM SIGGRAPH Symposium on Geometry Processing*, pages 156–164, 2003.
22. T. Ke, M. Maire, and S. X. Yu. Multigrid neural architectures. In *CVPR*, 2017.

23. D. P. Kingma and M. Welling. Auto-encoding variational bayes. *CoRR*, abs/1312.6114, 2013.
24. R. Klokov and V. Lempitsky. Escape from cells: Deep kd-networks for the recognition of 3d point cloud models. In *ICCV*, 2017.
25. Y. Li, S. Pirk, H. Su, C. R. Qi, and L. J. Guibas. Fpnn: Field probing neural networks for 3d data. In *NIPS*, 2016.
26. C.-H. Lin, C. Kong, and S. Lucey. Learning efficient point cloud generation for dense 3d object reconstruction. In *AAAI Conference on Artificial Intelligence (AAAI)*, 2018.
27. T.-Y. Lin, P. Dollár, R. Girshick, K. He, B. Hariharan, and S. Belongie. Feature pyramid networks for object detection. In *CVPR*, 2017.
28. Z. Lun, M. Gadelha, E. Kalogerakis, S. Maji, and R. Wang. 3d shape reconstruction from sketches via multi-view convolutional networks. In *International Conference on 3D Vision (3DV)*, 2017.
29. J. Masci, D. Boscaini, M. M. Bronstein, and P. Vandergheynst. Geodesic convolutional neural networks on riemannian manifolds. arXiv preprint arXiv:1501.06297, 2015.
30. D. Maturana and S. Scherer. Voxnet: A 3d convolutional neural network for real-time object recognition. In *IROS*, pages 922–928, 2015.
31. C. R. Qi, H. Su, M. Nießner, A. Dai, M. Yan, and L. Guibas. Volumetric and multi-view cnns for object classification on 3d data. In *Computer Vision and Pattern Recognition (CVPR)*, 2016.
32. C. R. Qi, L. Yi, H. Su, and L. J. Guibas. Pointnet++: Deep hierarchical feature learning on point sets in a metric space. In *NIPS*, 2017.
33. G. Riegler, A. O. Ulusoy, and A. Geiger. Octnet: Learning deep 3d representations at high resolutions. In *Conference on Computer Vision and Pattern Recognition*.
34. A. Sharma, O. Grau, and M. Fritz. Vconv-dae: Deep volumetric shape learning without object labels. In *ECCV Workshops*, 2016.
35. M. Simonovsky and N. Komodakis. Dynamic edge-conditioned filters in convolutional neural networks on graphs. In *CVPR*, 2017.
36. K. Simonyan and A. Zisserman. Very deep convolutional networks for large-scale image recognition. arXiv preprint arXiv:1409.1556, 2014.
37. A. A. Soltani, H. Huang, J. Wu, T. Kulkarni, and J. Tenenbaum. Synthesizing 3d shapes via modeling multi-view depth maps and silhouettes with deep generative networks. In *CVPR*, 2017.
38. H. Su, V. Jampani, D. Su, S. Maji, E. Kalogerakis, M.-H. Yang, and J. Kautz. SPLATNet: Sparse lattice networks for point cloud processing. *arXiv preprint arXiv:1802.08275*, 2018.
39. H. Su, S. Maji, E. Kalogerakis, and E. G. Learned-Miller. Multi-view convolutional neural networks for 3d shape recognition. In *ICCV*, 2015.
40. H. Su, C. Qi, K. Mo, and L. Guibas. Pointnet: Deep learning on point sets for 3d classification and segmentation. In *CVPR*, 2017.
41. M. Tatarchenko, A. Dosovitskiy, and T. Brox. Octree generating networks: Efficient convolutional architectures for high-resolution 3d outputs. In *IEEE International Conference on Computer Vision (ICCV)*, 2017.
42. P.-S. Wang, Y. Liu, Y.-X. Guo, C.-Y. Sun, and X. Tong. O-CNN: Octree-based convolutional neural networks for 3d shape analysis. *ACM Transactions on Graphics (SIGGRAPH)*, 36(4), 2017.
43. J. Wu, C. Zhang, T. Xue, W. T. Freeman, and J. B. Tenenbaum. Learning a probabilistic latent space of object shapes via 3d generative-adversarial modeling.
44. Z. Wu, S. Song, A. Khosla, F. Yu, L. Zhang, X. Tang, and J. Xiao. 3D Shapenets: A deep representation for volumetric shapes. In *Conference on Computer Vision and Pattern Recognition (CVPR)*, 2015.
45. L. Yi, H. Su, X. Guo, and L. Guibas. SyncSpecCNN: synchronized spectral cnn for 3d shape segmentation. In *CVPR*, 2017.

46. F. Yu and V. Koltun. Multi-scale context aggregation by dilated convolutions. *arXiv preprint arXiv:1511.07122*, 2015.

# Supplemental Material – Multiresolution Tree Networks for 3D Point Cloud Processing

Matheus Gadelha, Rui Wang and Subhransu Maji

College of Information and Computer Sciences,
University of Massachusetts - Amherst
{mgadelha, ruiwang, smaji}@cs.umass.edu

## 1 Shape segmentation model

As mentioned in the paper, MRTNet can also be applied to shape part segmentation. Given an input shape represented as a point cloud, the task is to segment the shape into meaningful parts. For example, a chair shape may be segmented into back, seat, leg parts etc. Here we describe the segmentation model and experimental results. The segmentation model follows a similar network to the MR-VAE, with two main differences:

1. We add skip-connections, similar to what's used in UNets [3] for image segmentations. Specifically, each intermediate tensor in the MR-Encoder is concatenated to the tensor of the same size in the MR-Decoder.
2. The output point cloud dimensionality is changed to 50 (instead of 3), i.e. each point is described by a part classification score for each of the 50 total possible parts covering the 16 object categories in ShapeNet.

To spatially sort the input point cloud, we use the RPtree, which according to our experiments leads to slight improvement for segmentation than using KDtree. Specifically, the splitting axes are precomputed random vectors sampled uniformly over the unit sphere. The same set of splitting axes are used for all shapes. In other words, it's almost the same as the vanilla KDtree except the axes are chosen as random vectors instead of $xyz$. This ensures a consistent ordering and relationship between neighboring points, which facilitate a dense classification task like part segmentation.

**Experiments.** We trained our segmentation model described above on the ShapeNet part segmentation benchmark, which contains groundtruth part labels for 16,881 shapes covering 16 different categories annotaded with 50 parts in total. Each object has 2 to 6 parts. Since this dataset is already represented as point clouds, it is not necessary to perform any additional point sampling. However, each shape may contain a varying number of points, so for each shape we randomly duplicate existing points until we have 4096 points. We follow the evaluation protocol used in prior work [2,4], where labels that do not appear in a particular category are ignored during evaluation. In other words, only the predictions corresponding to labels of a particular category are used.

Similar to the classification network, the segmentation network is trained by minimizing a cross-entropy using an Adam optimizer with learning rate $10^{-3}$ and $\beta = 0.9$. The learning rate is also divided by 2 every 5 training epochs. The first three convolutional layer have 32 filters and the next three have 64. We also apply test-time anisotropic scale augmentation. The scaling factors are sampled uniformly at random from the interval [0.8, 1.2]. At inference time, we compute 16 different scaled versions of the point cloud and return the mean classification score for each point.

| | #instances | 2690 | 76 | 55 | 898 | 3758 | 69 | 787 | 392 | 1547 | 451 | 202 | 184 | 283 | 66 | 152 | 5271 |
|---|---|---|---|---|---|---|---|---|---|---|---|---|---|---|---|---|---|
| | mean class | mean object | air- planes | bag | cap | car | chair | ear- phone | guitar | knife | lamp | laptop | motor- cycle | mug | pistol | rocket | skate board | table |
| MRTNet | 79.3 | 83.0 | 81.0 | 76.7 | 87.0 | 73.8 | 89.1 | 67.6 | 90.6 | 85.4 | 80.6 | 95.1 | 64.4 | 91.8 | 79.7 | 57.0 | 69.1 | 80.6 |
| KDNet [2] | 77.4 | 82.3 | 80.1 | 74.6 | 74.3 | 70.3 | 88.6 | 73.5 | 90.2 | 87.2 | 81.0 | 94.9 | 57.4 | 86.7 | 78.1 | 51.8 | 69.9 | 80.3 |
| PointNet [4] | 80.4 | 83.7 | 83.4 | 78.7 | 82.5 | 74.9 | 89.6 | 73.0 | 91.5 | 85.9 | 80.8 | 95.3 | 65.2 | 93.0 | 81.2 | 57.9 | 72.8 | 80.6 |
| 3DCNN [4] | 74.9 | 79.4 | 75.1 | 72.8 | 73.3 | 70.0 | 87.2 | 63.5 | 88.4 | 79.6 | 74.4 | 93.9 | 58.7 | 91.8 | 76.4 | 51.2 | 65.3 | 77.1 |
| Yi, 2016 [5] | 79.0 | 81.4 | 81.0 | 78.4 | 77.7 | 75.7 | 87.6 | 61.9 | 92.0 | 85.4 | 82.5 | 95.7 | 70.6 | 91.9 | 85.9 | 53.1 | 69.8 | 75.3 |

**Table 1**: Shape segmentation results. Numbers reported here are the mean intersection over union (mIoU) scores. The table shows comparisons between methods that use 3D position information without normal information.

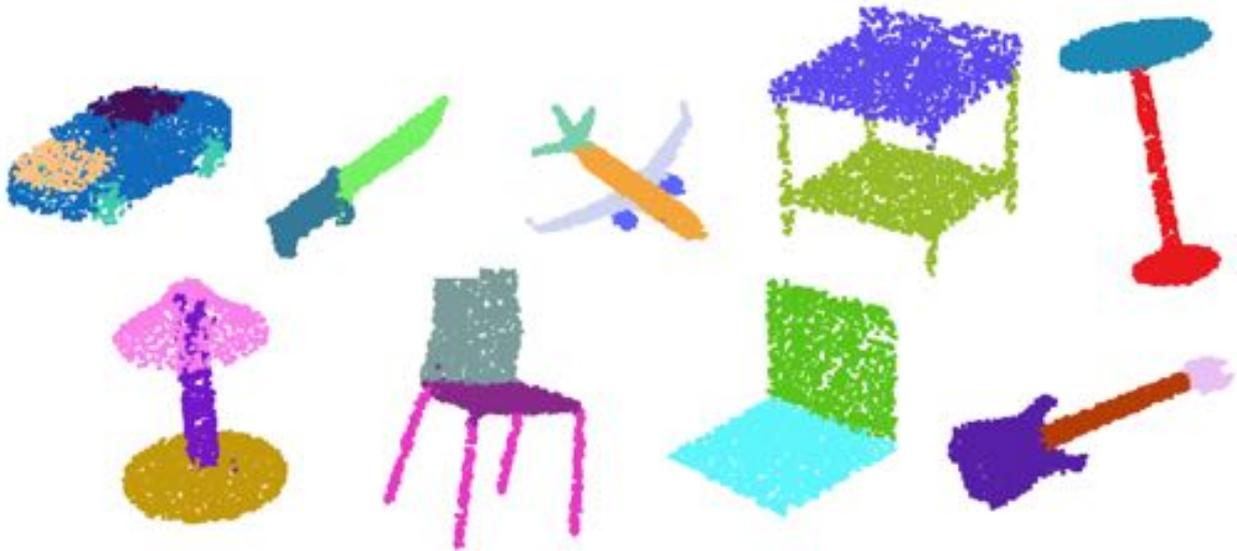

**Fig. 1:** Qualitative results for the shape segmentation task. Each different segmentation part is shown in a different color.

Figure 1 shows qualitative results from our part segmentation, and Table 1 lists the mean intersection over union (mIoU) results for all 16 categories. Our approach produces competitive results in comparison with the state of the art. In particular, it outperforms the recently proposed KDNet by nearly 2% in the mean per category mIoU. However, it lacks behind some other recent works, like PointNet [4].

In comparison to the multiresolution UNets, the single resolution counterpart leads to a drop in about 1%. We believe that the drop here is much smaller in comparison to the classification task because the multiresolution representation is replicating similar benefits to UNet, where information from different scales is added directly to the decoder. In order to verify this hypothesis we trained a segmentation network without the skip connections from UNet. The multiresolution model obtained an accuracy of 79.82%, while the baseline had 76.14%, which is 3% lower.

Increasing filter size in the single resolution network did not yield any benefits. For example, increasing the kernel size from 2 to 8, led to a drop of about 1% in performance. This demonstrates that the multiresolution network adds more than simply increasing the receptive fields.

## 2 Image-to-Shape Inference: Additional Results

In this section, we show additional results of image-to-shape inference using MRTNet.

**Reconstructed meshes**. For some applications, such as 3D printing, the output is required to be triangle meshes instead of just point clouds. To do so, we take the image-to-shape inference results (each shape containing 4K points), create a $0.025^3$ cube centered at each point, then apply Poisson Surface Reconstruction to create triangle meshes. This is a simple way to create a mesh from a point cloud, without normal estimation (due to the relatively low point count). Figure 2 shows renderings of the reconsturcted meshes. The input images are Internet photos from Figure 6 in the paper. We rendered each mesh in wireframe mode to reveal the underlying triangle meshes. Some geometric details in the point clouds are necessarily smoothed out due to surface reconsturctions. Nonetheless, the reconsturcted meshes are reasonably faithful to the input images.

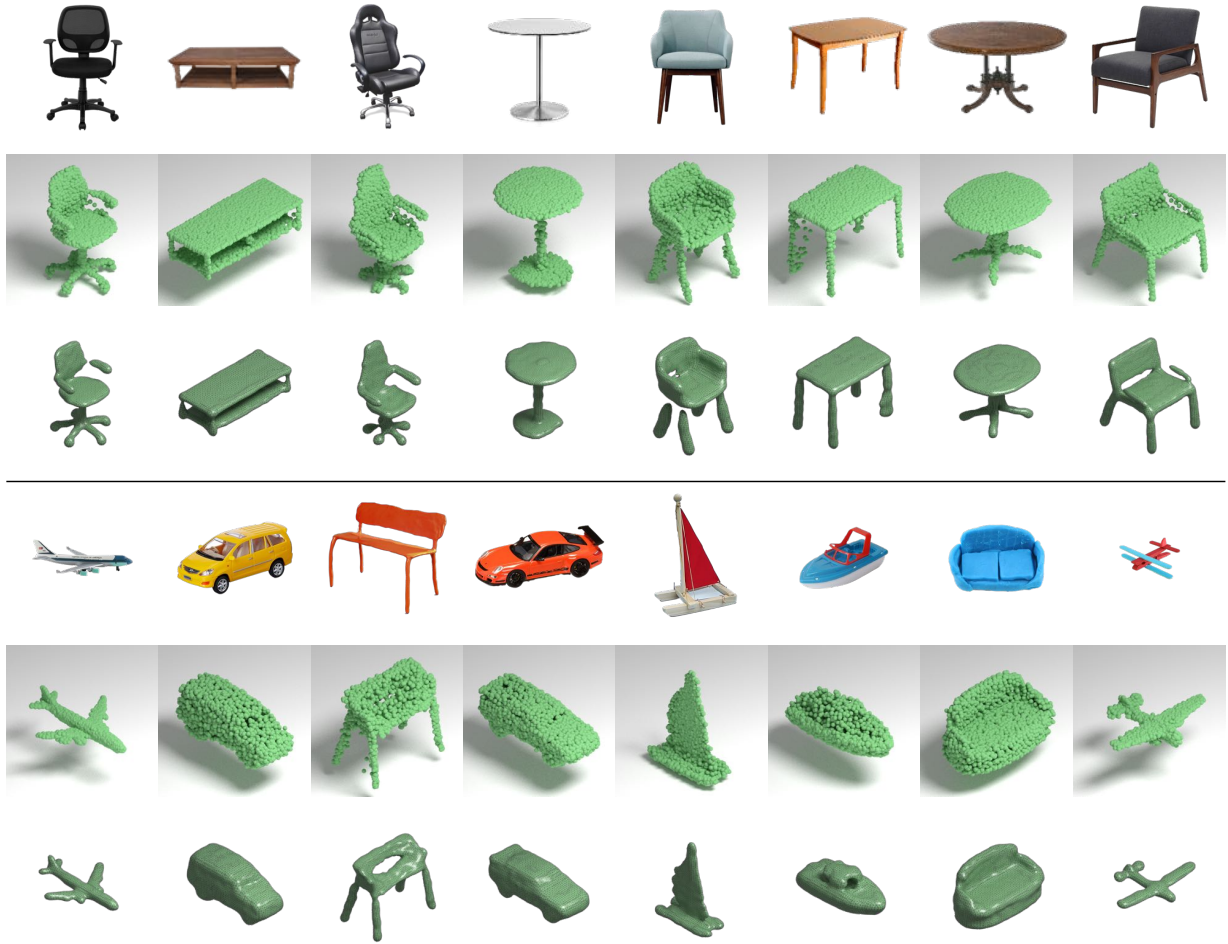

**Fig. 2:** Reconsturcted meshes from point clouds generated by applying MRTNet on Inernet photos of furnitures and toys. From top-down, the first image in each example is the input photo, the second is the point cloud (each 4K points) generated by MRTNet, the third is a rendering of the reconstructed mesh. We rendered each mesh in wireframe mode to reveal the underlying triangles. Zoom in for details.

**Full ablation studies.** In the paper, for image-to-shape inference experiments, we presented a summary of ablation studies of MRTNet in comparison with a single-resolution baseline, and a fully-connected baseline. Here we present the full ablation study results, covering all 13 shape categories. Refer to Table 2. Note that MRTNet is consistently better than both baselines.

## 3  Unsupervised Shape Generation (MR-VAE): Additional Results

In this section, we show additional results of unsupervised shape generation.

| Category | MRTNet | Single Res. | Fully Connected |
|---|---|---|---|
| airplane | **0.976 / 0.920** | 1.142 / 1.185 | 1.258 / 1.423 |
| bench | **1.438 / 1.326** | 1.616 / 1.666 | 1.753 / 2.358 |
| cabinet | **1.774 / 1.602** | 1.913 / 1.916 | 1.980 / 2.263 |
| car | **1.395 / 1.303** | 1.511 / 1.476 | 1.583 / 1.668 |
| chair | **1.650 / 1.603** | 1.789 / 1.927 | 1.982 / 2.385 |
| display | **1.815 / 1.901** | 2.060 / 2.301 | 2.185 / 3.029 |
| lamp | **1.944 / 2.089** | 1.953 / 2.608 | 2.163 / 3.400 |
| speaker | **2.165 / 2.121** | 2.336 / 2.456 | 2.374 / 2.913 |
| rifle | **1.029 / 1.028** | 1.191 / 1.213 | 1.291 / 1.402 |
| sofa | **1.768 / 1.756** | 1.885 / 2.022 | 1.985 / 2.421 |
| table | **1.570 / 1.405** | 1.689 / 1.562 | 1.808 / 2.049 |
| telephone | **1.346 / 1.332** | 1.637 / 1.677 | 1.643 / 2.342 |
| watercraft | **1.394 / 1.490** | 1.482 / 1.792 | 1.703 / 2.202 |
| **mean** | **1.559 / 1.529** | 1.708 / 1.831 | 1.824 / 2.297 |

**Table 2: Single-image shape inference.** Full results of the ablation studies covering all 13 categories. Note that MRTNet is consistently better than both baselines.

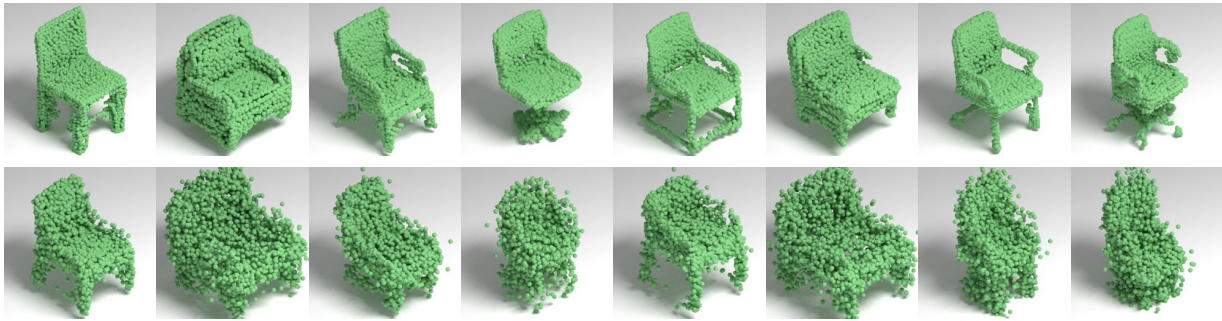

**Fig. 3:** Chairs generated by randomly sampling the encoding. Top: results from our MR-VAE; Bottom: results by using a fully connected (FC) baseline.

**Comparison with FC Decoder.** We experimented using fully-connected (FC) decoders as a baseline for MR-VAE. The generated shapes have a much lower quality: see results in Fig. 3. This is not surprising as there is no prior work that employs these types of decoders alone. Solutions like [1] use FC layers, but as an additional branch to a much deeper convolutional architecture.

Besides, generating shapes by sampling from a random noise is much harder than reconstructing input shapes. The shapes presented in Fig. 3 are all **sampled** from a random noise and **not** reconstructions (which would be easy to generate at high quality). Moreover, we are generating shapes with 4K points, which is more difficult for a vanilla decoder, because it tends to generate a lot of misplaced points, as we can see in Fig. 3 bottom row. Similarly, single resolution decoders produce low-quality samples as seen in Fig. 7 bottom row (in the main paper).

Finally, note that PointNet and Kd-net cannot be used as **decoders** since the former ignores the ordering of points while the later conditions the processing on the splits.

**Visualization of MR-VAE encodings using t-SNE.** In Figure 4 we show visualizations of encodings learned by MR-VAE using t-SNE. Specifically, we randomly selected 1000 shapes from the ShapeNet dataset, computed their encodings learned by MR-VAE, then applied t-SNE to compute the 2D coordinate of each encoding, and fi-nally rendered all 1000 shapes on a 2D plane. From the results, we observe that similar shapes tend to stay together, indicating the ability of MR-VAE on learning the latent representations of the shapes.

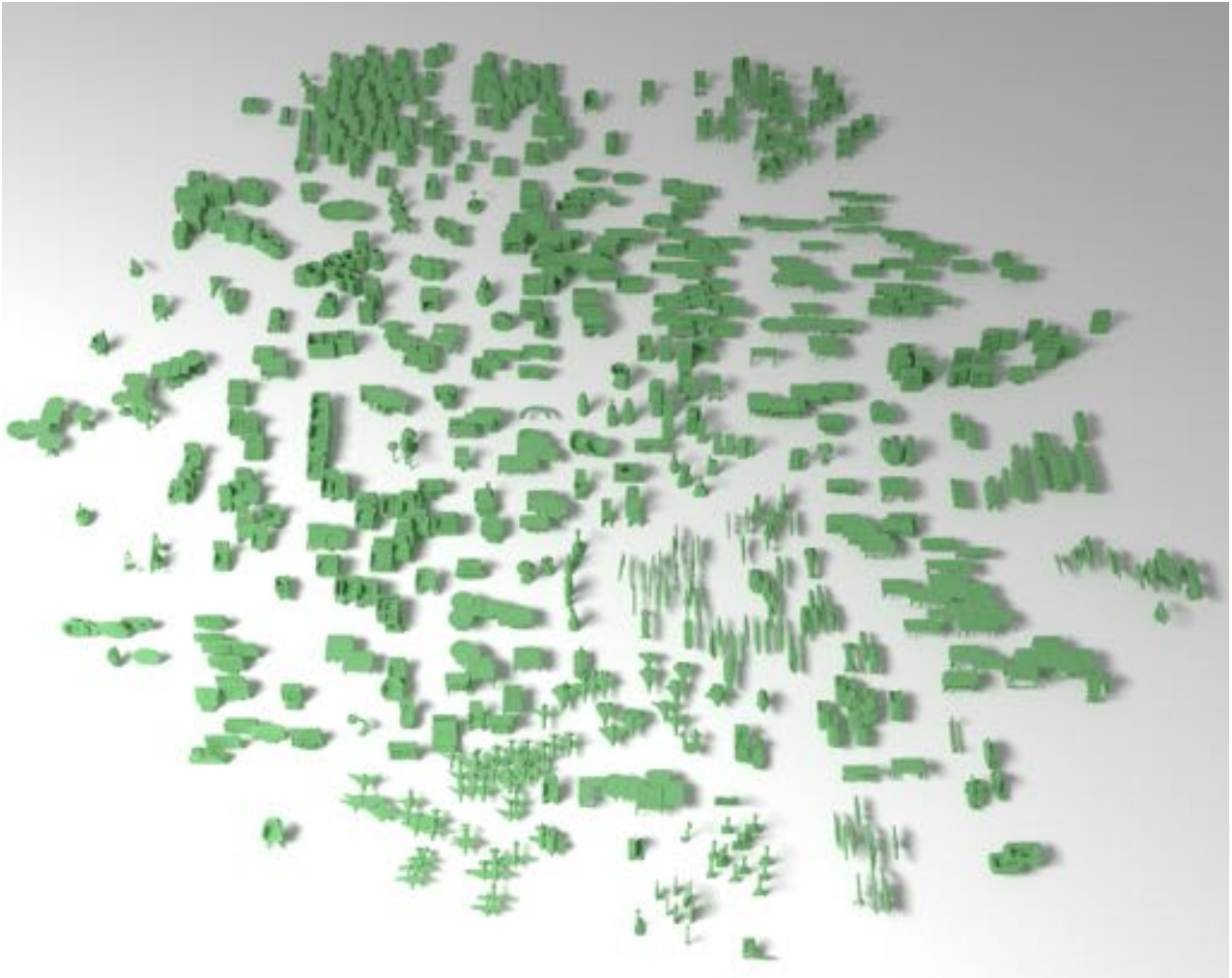

**Fig. 4: ShapeNet shapes arranged according to their encodings learned by MR-VAE.** 1000 samples from our model trained in the ShapeNet data. The position of the models in the plane is computed after running t-SNE on the latent representation of the shapes. Zoom in for details.

# References

1. H. Fan, H. Su, and L. Guibas. A point set generation network for 3d object reconstruction from a single image. In *IEEE Conference on Computer Vision and Pattern Recognition*, 2017.
2. R. Klokov and V. Lempitsky. Escape from cells: Deep kd-networks for the recognition of 3d point cloud models. In *ICCV*, 2017.
3. O. Ronneberger, P.Fischer, and T. Brox. U-net: Convolutional networks for biomedical image segmentation. In *Medical Image Computing and Computer-Assisted Intervention (MICCAI)*, LNCS, pages 234–241, 2015.
4. H. Su, C. Qi, K. Mo, and L. Guibas. Pointnet: Deep learning on point sets for 3d classification and segmentation. In *CVPR*, 2017.
5. L. Yi, V. G. Kim, D. Ceylan, I.-C. Shen, M. Yan, H. Su, C. Lu, Q. Huang, A. Sheffer, and L. Guibas. A scalable active framework for region annotation in 3d shape collections. *SIGGRAPH Asia*, 2016.